# Accurate Tennis Court Line Detection on Amateur Recorded Matches

Sameer Agrawal[a], Ragoth Sundararajan[a], Vishak Sagar[a]

[a]NicheSolv India Private Limited, 129, 3rd floor, 8th Main Rd, 3rd Phase, J. P. Nagar, Bengaluru, Karnataka 560078, India

## ABSTRACT

Typically, tennis court line detection is done by running Hough-Line-Detection to find straight lines in the image, and then computing a transformation matrix from the detected lines to create the final court structure. We propose numerous improvements and enhancements to this algorithm, including using pretrained State-of-the-Art shadow-removal and object-detection ML models to make our line-detection more robust. Compared to the original algorithm, our method can accurately detect lines on amateur, dirty courts. When combined with a robust ball-tracking system, our method will enable accurate, automatic refereeing for amateur and professional tennis matches alike.

**Keywords:** Tennis court line detection, AI referee, Machine learning, Automatic sports analysis, AI based shadow removal, YOLO object detection, Image and video processing, Computer vision

## 1. INTRODUCTION

Automatic tennis-court-line detection is an important step towards creating tennis-match analysis software. Although important in many sports, court-line detection is an especially salient problem in tennis due to the importance of in vs out-of-bounds balls. Tennis courts are also especially well-suited for this problem because they are relatively smaller in size compared to other courts, allowing most if not all of the court to be visible in a typical tennis match video.

### 1.1. Challenges to automatic court line detection

However, there are numerous challenges to automatic court-line detection. The ideal court for this task would be recorded from a relatively high angle in which all important points on the court are clearly visible, as shown in Figure 1a. However, many courts, especially in amateur matches, are recorded from a lower angle (Figure 1b) in which the far side of the court is not visible, or the lines may be partially covered by another part of the video (Figure 1c), or the court may be scuffed so that the lines themselves are not clearly visible (Figure 1d). All of these present obstacles to any court-line detection model, especially when dealing with non-professional matches, and must therefore be dealt with.

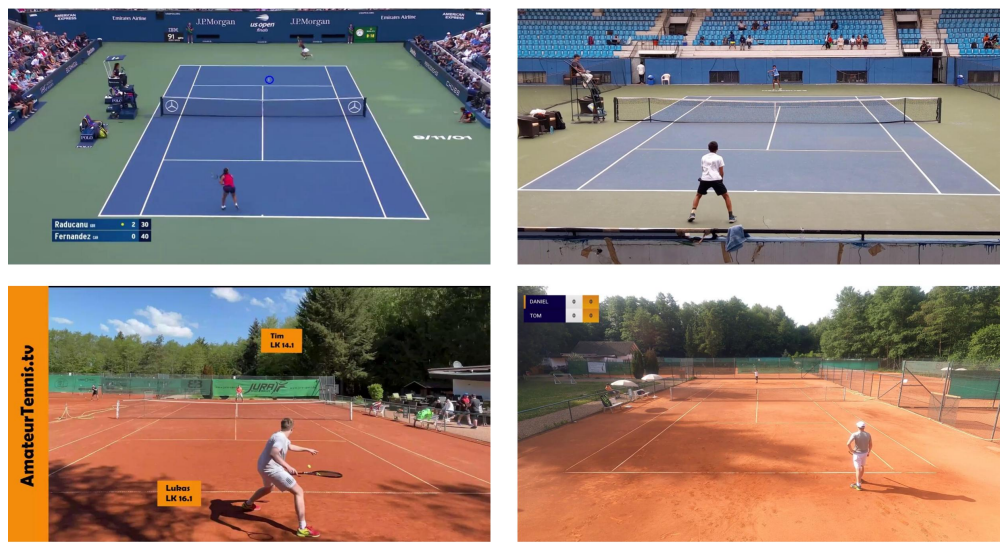

Figure 1. Examples of tennis courts. *Top left:* A "perfect" court. *Top Right:* Low camera angle. *Bottom Left:* court is obscured. *Bottom Right: S*cuffed court lines.

### 1.2. Our contributions

- To allow our algorithm to handle extremely bright courts where the white lines are not clearly visible such as that in Figure 1d, a filtration algorithm based on detection of the court's color is proposed, allowing pixels identified to be part of the court to be removed entirely from the image.

- To handle instances in which a low camera angle makes the far side of the court practically impossible to see, such as in Figures 1b and 1d, the image is first cropped to include only the near side, using YOLO-detected net coordinates. This detected half-court is then extended to the full image.

- To handle courts in which shadows obscure some of the field, such as in Figures 1c and 1d, advanced Shadow-Removal models are used to create an image which the algorithm can better parse through.

## 2. RELATED WORK

### 2.1. Homography estimation for sports-field registration

The algorithm which we improve upon in this paper is a homography estimation algorithm. This algorithm first performs a filtration process on the court to make the court lines more visible. It then runs Hough Line Detection on this filtered image and calculates all the intersections between the detected lines. It then uses these intersection points to estimate a transformation matrix between a reference tennis court diagram and the tennis court in the frame[1].

### 2.2. Introduction of AI models

In the past, Deep Learning models have also been introduced to the problem[2,3]. Neural Networks can be used to optimize the initial homography matrix via the Gradient Descent algorithm[2]. In other methods, a homography matrix is not calculated, and instead a CNN is used to generate a mask highlighting important areas of the court[3].

### 2.3. Key differences of our approach

The method described in Section 2.1, although groundbreaking, is not optimized for non-professional recordings specifically, and is unable to handle amateur courts. Meanwhile, ML-based methods are limited by the data they were trained on, which favors professional, televised matches. By comparison, our method uses pre-trained models such as YOLO and ShadowFormer, which leads to a more robust algorithm capable of handling amateur courts.

## 3. OUR WORK

Our work improves upon the homography-estimation algorithm by incorporating the following new approaches: First, we use a shadow-removal model to make court lines obscured by shadow more clear. Second, we use a tennis-net detection model to crop the frame to include only the near-side of a court. Third, we have modified the pixel-by-pixel filtering algorithm to use the court color. Fourth, we use a scoring algorithm to determine the best court-detection out on a set of detected courts. These improvements are described in more detail below.

### 3.1. Shadow removal

We make use of two state-of-the-art shadow removal models: the MTMT model for shadow-detection[4], and the ShadowFormer model for shadow-removal[5]. The MTMT model is used to generate a shadow-mask, and this shadow-mask, along with the original image, is passed into ShadowFormer for final shadow removal. An example of this in action on a real tennis-court frame can be seen in Figure 2.

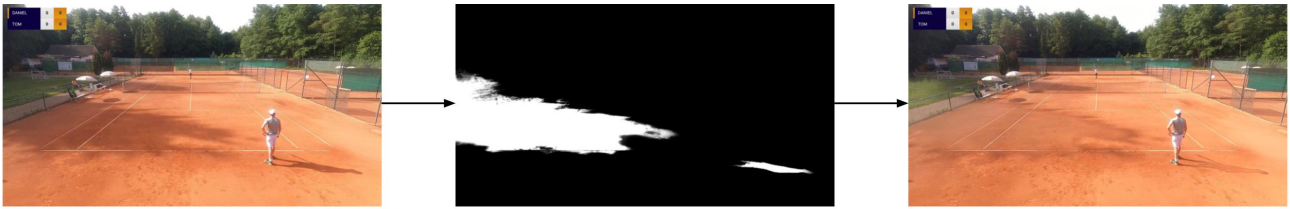

Figure 2. Our shadow-removal process. The original image is passed into a Multi-task Mean Teacher (MTMT) model, which generates a shadow-mask. Both images are then passed to Shadow-Former, which performs final shadow removal.

## 3.2. YOLO-v5 net detection

After shadows are removed, a YOLO-v5 object detection model is used to detect the tennis net. The detected net coordinates are then used to crop the image to only include the near-side of the court, as shown in Figure 3. By only running court detection on the near side, and then simply extending the detected court to the far side, the algorithm is made more accurate and faster.

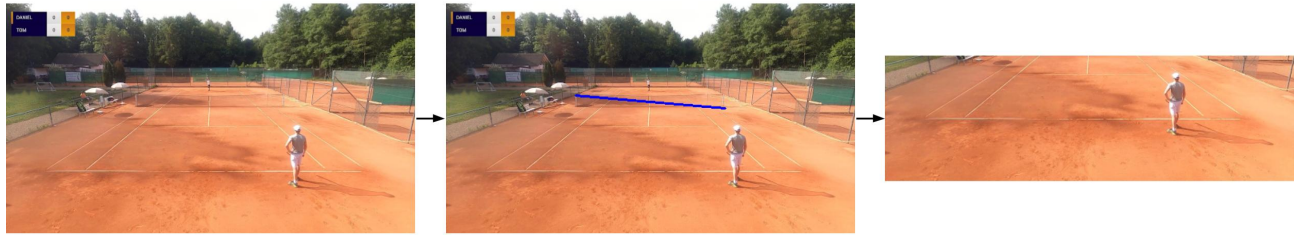

Figure 3. Example of image cropping using YOLO net detection. *Left: O*riginal, shadow-removed image; *Middle:* Diagonal line representing YOLO-detected net coordinates; *Right:* Final cropped image.

## 3.3. Pixel-by-pixel filtering

Previous methods for image filtering, such as that described in Section 2.1, use a simple pixel threshold to filter out any dark pixels in the image, thus theoretically only leaving the bright white lines behind. However, the ideal threshold value can change drastically from image to image, and is therefore not a robust method, as shown in Figure 4.

By comparison, we propose a filtering algorithm based on finding the court color. The steps of the algorithm are:
1. Place 1000 points randomly on the image.
2. Find the color of each point, and then return the color shared by the most points. (Figure 5, middle column*)*
3. Loop through the image pixel-by-pixel, and at a given pixel *A*, look in a 7x7 square around *A*. If at least 4 pixels within this square match the detected court color, AND pixel *A* itself does NOT match the detected court color, then pixel *A* is detected to be part of a court line and is white. (Figure 5, right column)

Figure 5 shows this new filtration algorithm in action on the same two courts as in Figure 4. The new filtration algorithm is more capable of generalizing to a variety of courts compared to the original threshold-based filtration algorithm.

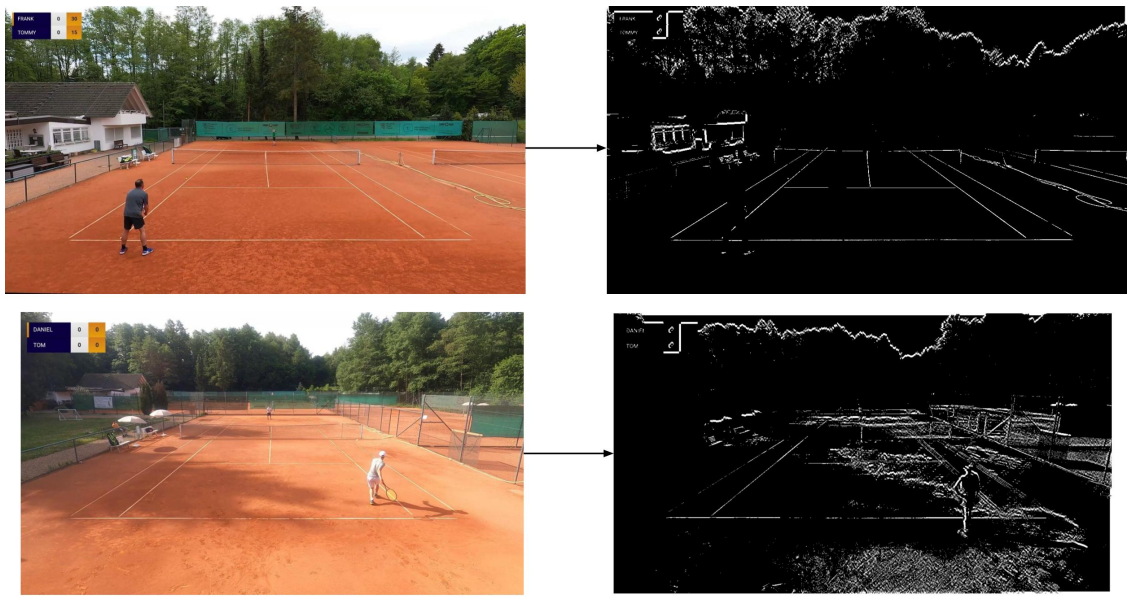

Figure 4. The old pixel filtration algorithm. A threshold value of 180 works well for the top image, but is too low for the bottom image, as it leaves behind residual white pixels throughout the court.

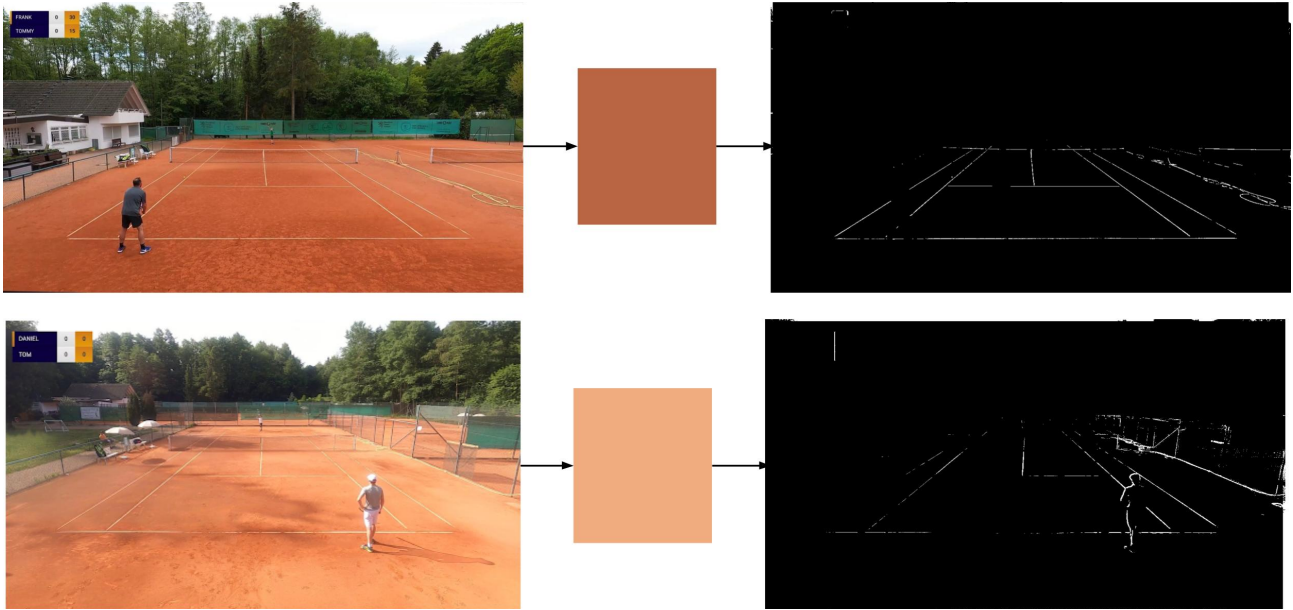

Figure 5. The new filtration algorithm. *Left:* Original Image; *Middle:* Detected court color; *Right:* Final filtered image

### 3.4. Scoring algorithm

To ensure accurate court detections, the algorithm is run on a *video*, as opposed to individual images. The detection process is run on randomly selected frames from the video, and when a set of court-detections is obtained, a scoring algorithm is used to find the most accurate detection. The algorithm works as follows:
1. The court detection is drawn onto the image.
2. Given both this drawn image and the original image, the number of drawn pixels corresponding to bright white pixels in the original image are counted. This is the final score.

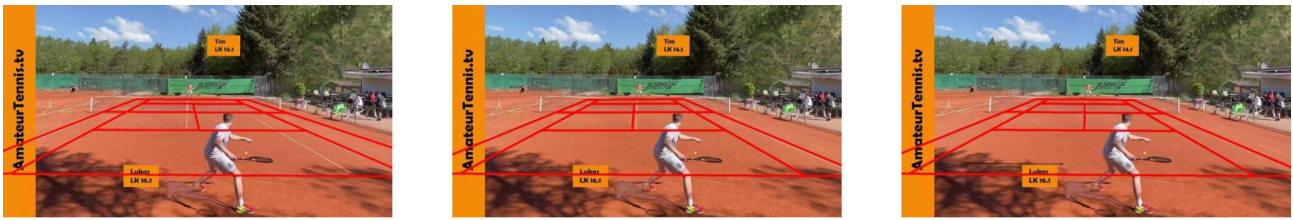

Figure 6. Three court detections of varying scores. *Left:* Score of 356; *Middle:* 812; *Right:* 844. A higher score represents a more accurate detection.

## 4. RESULTS

We compared our method with a version of the Homography-Estimation algorithm described in Section 2.1, which has been designated as the "Old Method" in the table. We used 180 as the threshold value. In order to draw a more accurate and fair comparison between the old method and our method, we do not run our algorithm on a video, and instead, only run it on individual frames. The results of our experiments are in Table 1.

### 4.1. Dataset

There exist no compiled datasets of non-professional matches. So, we obtained our testing data by either recording our own tennis matches, or by taking videos from a website called AmateurTennis.tv, which features hundreds of amateur tennis match videos. In this paper, we present the results from 7 different videos, four of which are from amateur matches and three from professional matches.

### 4.2. Evaluation metrics

To determine if a given court prediction is accurate, ground truth court lines were made for each video. If the score (described in Section 3.4) of a detected court was within 5% of the score of the ground truth court lines, then the given court detection was marked as correct.

Table 1. Results of our experiments.

| Court Number | # of Frames Tested On | Accuracy - Old Method | Accuracy - New Method |
| --- | --- | --- | --- |
| Court #1 | 159 (new); 15 (old) | 60.0% | **84.3%** |
| Court #2 | 105 | 79.1% | **86.7%** |
| Court #3 | 123 | 92.7% | **100.0%** |
| Court #4 | 112 | 0.0% | **81.3%** |
| Court #5 | 50 | **96.0%** | 86.0% |
| Court #6 | 51 | 86.3% | **88.3%** |
| Court #7 | 65 | 50.8% | **100.0%** |

### 4.3. Analysis

Matches 1-4 are non-professional, and our method uniformly performs better on these. The old method was limited because it frequently detected extraneous lines due to its filtering algorithm. Our method's filtration algorithm was generally able to filter out non-court lines far better, leading to both more accurate and faster detections. Court 5 was a Wimbledon court with alternating bands of color, which led to confusion in the detected court-color.

## 5. CONCLUSION AND FUTURE WORK

We presented numerous enhancements and improvements to an existing court-detection framework, giving our method increased accuracy across the board, but especially on amateur and non-professional courts.

We plan to further explore new filtration algorithms, including incorporating a CNN to extract and highlight features of a court. We also plan to investigate other net-detection and shadow-removal models for increased speed and accuracy.